\documentclass[acmtog]{acmart}

\usepackage{colortbl}  
\usepackage{microtype}
\usepackage{graphicx}
\usepackage{subfigure}
\usepackage{booktabs} 
\usepackage{hyperref}
\usepackage{bm} 

\usepackage{amsmath}
\usepackage{mathtools}
\usepackage{amsthm}
\usepackage{amssymb}
\usepackage{multirow}
\usepackage{enumitem}
\usepackage{tabularx}   
\usepackage{dblfloatfix}

\theoremstyle{plain}

\theoremstyle{definition}

\theoremstyle{remark}

\newcommand{\modelname}{\textit{BlobCtrl}}
\newcommand{\benchmarkname}{\textit{BlobBench}}
\newcommand{\dataname}{\textit{BlobData}}

\copyrightyear{2025}
\acmYear{2025}
\setcopyright{acmlicensed}\acmConference[SA Conference Papers '25]{SIGGRAPH Asia 2025 Conference Papers}{December 15--18, 2025}{Hong Kong, Hong Kong}
\acmBooktitle{SIGGRAPH Asia 2025 Conference Papers (SA Conference Papers '25), December 15--18, 2025, Hong Kong, Hong Kong}
\acmDOI{10.1145/3757377.3763897}
\acmISBN{979-8-4007-2137-3/2025/12}

\begin{document}

\title{BlobCtrl: Taming Controllable Blob for Element-level Image Editing}

\author{Yaowei Li}
\email{liyaowei01@gmail.com}
\affiliation{%
  \institution{SECE, Peking University}
  \country{China}
}

\author{Lingen Li}
\affiliation{%
  \institution{The Chinese University of Hong Kong}
  \city{Hongkong}
  \country{China}
}
\email{lgli@link.cuhk.edu.hk}

\author{Zhaoyang Zhang}
\affiliation{%
  \institution{ARC Lab, Tencent}
  \country{China}}
\authornotemark[2]
\email{zhaoyangzhang@link.cuhk.edu.hk}

\author{Xiaoyu Li}
\affiliation{%
  \institution{ARC Lab, Tencent}
  \country{China}}
\authornote{Project Lead.}
\email{xliea@connect.ust.hk}

\author{Guangzhi Wang}
\affiliation{%
  \institution{ARC Lab, Tencent}
  \country{China}}
\email{guangzhi.wang@u.nus.edu}

\author{Hongxiang Li}
\affiliation{%
  \institution{The Hong Kong University of Science and Technology}
  \city{Hongkong}
  \country{China}}
\email{lihxxxxxx@gmail.com}

\author{Xiaodong Cun}
\affiliation{%
  \institution{GVC Lab, Great Bay University}
  \country{China}}
\email{vinthony@gmail.com}

\author{Ying Shan}
\affiliation{%
  \institution{ARC Lab, Tencent}
  \country{China}}
\email{yingsshan@tencent.com}

\author{Yuexian Zou}
\affiliation{%
  \institution{SECE, Peking University}
  \country{China}}
\authornote{Corresponding Author.}
\email{zouyx@pku.edu.cn}

\begin{abstract}
As user expectations for image editing continue to rise, the demand for flexible, fine-grained manipulation of specific visual elements presents a challenge for current diffusion-based methods.
In this work, we present \modelname{}, a framework for element-level image editing based on a probabilistic blob-based representation. Treating blobs as visual primitives, \modelname{} disentangles layout from appearance, affording fine-grained, controllable object-level elements manipulation.
Our key contributions are twofold: 1) an in-context dual-branch diffusion model that separates foreground and background processing, incorporating blob representations to explicitly decouple layout and appearance; and 2) a self-supervised disentangle-then-reconstruct training paradigm with an identity-preserving loss function, along with tailored strategies to efficiently leverage blob-image pairs.
To foster further research, we introduce \dataname{} for large-scale training, and \benchmarkname{}, a benchmark for systematic evaluation. Experimental results demonstrate that \modelname{} achieves state-of-the-art performance in a variety of element-level editing tasks—such as object addition, removal, scaling, and replacement—while maintaining computational efficiency.
\end{abstract}


\begin{teaserfigure}
    \centering
    \vspace{-8pt}
    \includegraphics[width=0.82\linewidth]{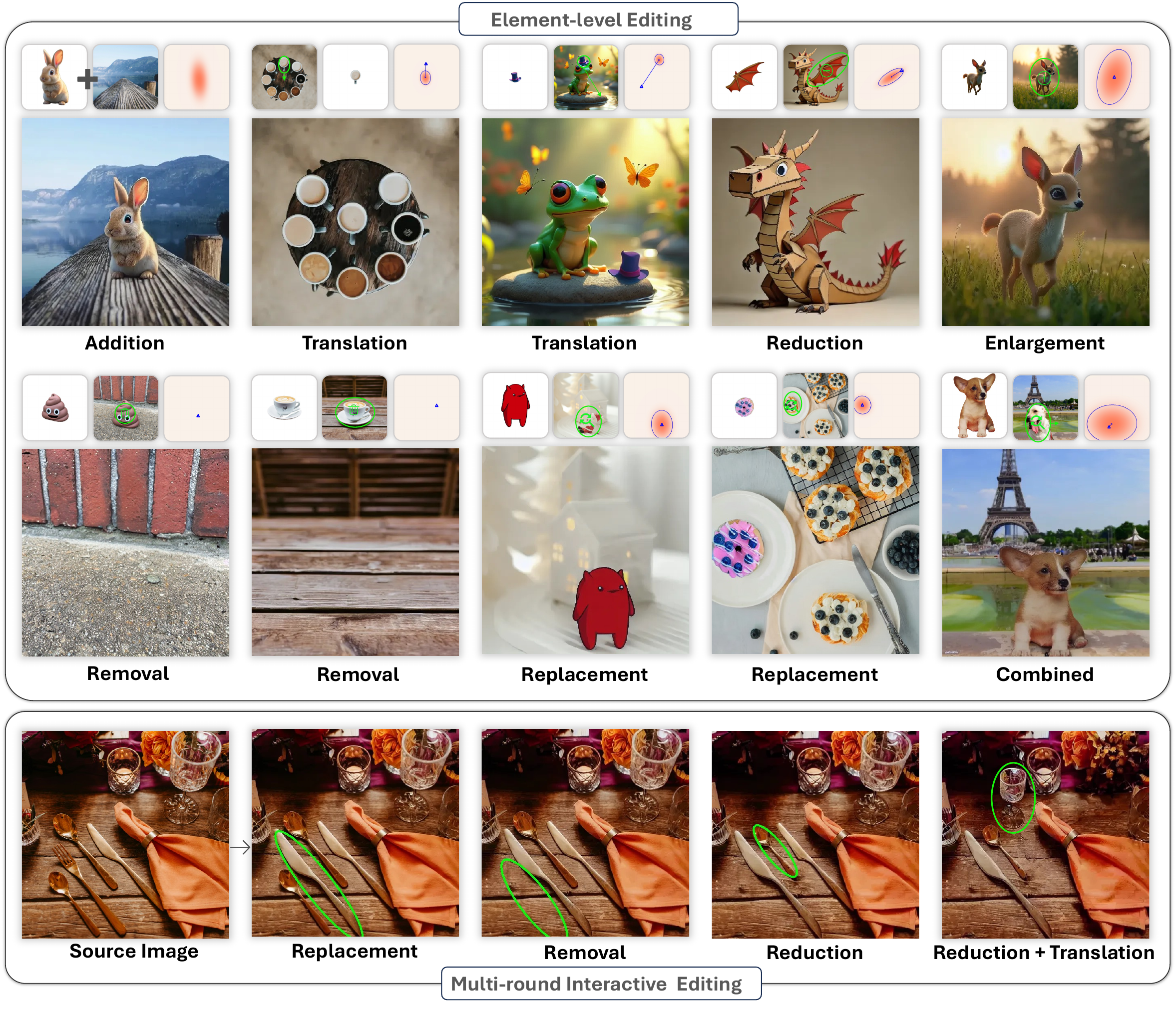}
    \vspace{-10pt}
    \caption{\modelname{} enables comprehensive element-level editing, supporting diverse operations such as addition, translation, scaling, removal, replacement, and their arbitrary combinations (top). Via iterative refinement, \modelname{} achieves precise, fine-grained control to realize the desired visual outcomes (bottom).}
    \label{fig:teaser}
\end{teaserfigure}


\maketitle

\section{Introduction}
\label{sec:introduction}
Element-level image editing aims to achieve fine-grained refinement of the layout and appearance of visual elements in existing images. While recent generative models\cite{ramesh2022hierarchical, flux2023, esser2024scaling, sheynin2024emu, shi2024seededit, yu2025anyedit} excel in high-quality image synthesis and editing, they often lack a straightforward approach for fine-grained control over individual visual elements.
Conventional controllable generative approaches~\cite{zhang2023adding, ye2023ip, li2023gligen, wang2024instantid} introduce spatial conditions (such as edge maps, bounding boxes) or identity conditions (like reference images or ID features) to generate new images from scratch. However, these methods cannot modify the layout and appearance of existing images, nor do they support interactive, multi-round, element-based editing operations such as visual element rearrangement.

Recent methods~\cite{zhang2023continuous, shi2023dragdiffusion, alzayer2024magic, mu2025editable, mao2025ace++, song2025insert, li2025ic} have explored fine-grained visual editing through optimization, segmentation, clustering, and drag-based approaches. However, these methods lack robust and flexible editing capabilities due to two main limitations: 1) undesirable changes in unedited regions during the editing process, and 2) reliance on video data for training, which leads to artifacts in edited content (e.g., failed inpainting of the original location when moving elements).

The essence of element-level visual representation lies in the flexible decoupling of layout and visual appearance. To this end, \modelname{} employs blobs as visual primitives to make the layout and appearance of the edited elements controllable. Formally, a blob is a probabilistic two-dimensional Gaussian distribution~\cite{carson1999blobworld}, and geometrically, it appears as an ellipse~\cite{niecompositional}. While prior works~\cite{niecompositional, epstein2022blobgan} use blobs to specify layouts for image synthesis from scratch, we further tame blobs to enable precise layout rearrangement and appearance replacement for fine-grained element-level editing, leveraging their 5-DoF (x, y, a, b, $\theta$) and opacity-aware operations to accurately control position, scale, and orientation.

We propose an in-context dual-branch diffusion architecture that decouples foreground and background processing using a blob-based representation. To better utilize blob-image pairs and avoid artifacts commonly seen in methods trained on video data, we introduce a self-supervised disentangle-then-reconstruct training paradigm with a carefully designed identity-preserving optimization objective.
Additionally, we introduce several tailored strategies: random data augmentation to prevent the model from falling into copy-paste local optima, and random feature dropout to enable more flexible diffusion inference. These design choices make \modelname{} an efficient, flexible solution for element-level image editing.

To scale up our method and ensure comprehensive evaluation, we introduce a new training dataset, \dataname{}, and a benchmark, \benchmarkname{}. Extensive quantitative and qualitative results demonstrate \modelname{}'s effectiveness in fine-grained element-level editing (addition, translation, scaling, removal, and replacement).

In a nutshell, our main contributions include:
\begin{itemize}[nosep, leftmargin=*]
    \item We propose \modelname{}, a novel approach that tames blobs as visual primitives to enable precise and flexible visual element editing, while effectively preserving their intrinsic characteristics.
    \item We introduce a self-supervised disentangle-then-reconstruct training paradigm with an identity-preserving loss function, along with tailored strategies to efficiently leverage blob-image pairs.
    \item We introduce \dataname{}, a comprehensive dataset specifically curated for training blob-based editing, alongside \benchmarkname{}, a rigorous benchmark for assessing element-level editing capabilities.
    \item Through extensive experimentation, we demonstrate that \modelname{} achieves superior performance compared to existing methods in element-level editing tasks, while maintaining computational efficiency and practical applicability.
\end{itemize}

 \begin{figure}[t]
    \centering
    \includegraphics[width=0.9\linewidth]{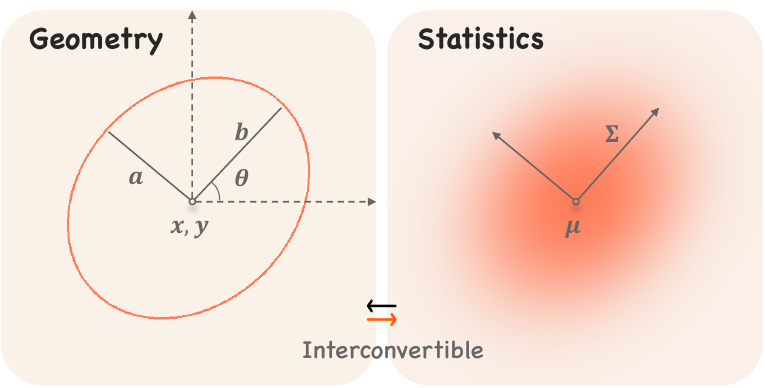}
    \vskip -0.1in
    \caption{\textbf{Blob Formula.} A blob can be represented in two equivalent forms: geometrically as an ellipse and statistically as a 2D Gaussian distribution.  The two forms are exactly equivalent and interchangeable.}
    \label{fig:illustration_blob}
    \vspace{-0.4cm}
\end{figure}

\section{Related Works}
\label{appendix:related_work}
\begin{figure*}[tbp]
\centering
\includegraphics[width=.98\textwidth]{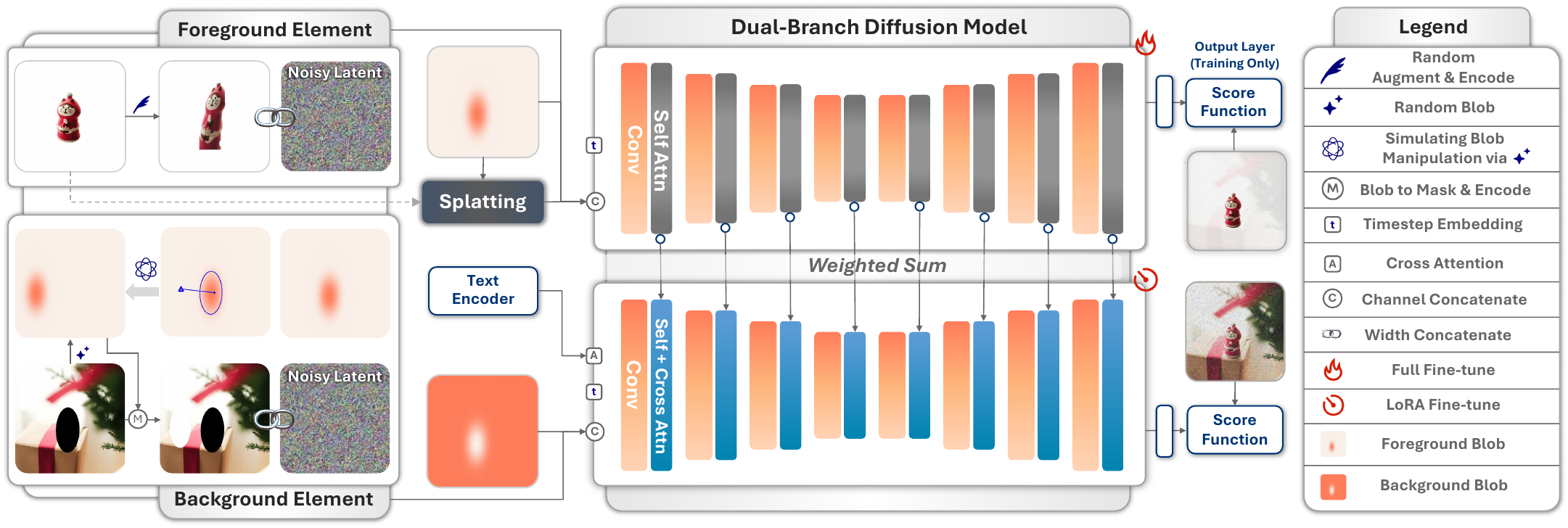}
\vspace{-2mm}
\caption{\textbf{Overview of \modelname{}.} Our framework employs a dual-branch architecture: a foreground branch for identity encoding and a background branch for scene preservation and fusion. Inputs are concatenated in an in-context manner (Sec.~\ref{sec:model_architecture}), and the model is trained using the proposed strategy (Sec.~\ref{sec:self-supervised}).}
\label{fig:framework}
\vspace{-4mm}
\end{figure*}

\paragraph{\textbf{Image Editing.}} Prompt-based image editing methods~\cite{hertz2023prompt, brooks2023instructpix2pix, huang2024smartedit, cao_2023_masactrl, li2024brushedit, shi2024seededit} primarily rely on text as editing instructions. Reference-based image editing methods~\cite{gal2022image, ruiz2023dreambooth, kumari2023multi, ye2023ip, wang2024instantid} focus on preserving the visual appearance of reference images in new scenarios. Most relevant to our work are spatial-based editing methods, which typically employ per-sample optimization algorithms~\cite{zhang2023continuous, yenphraphai2024image}, point-based drag methods~\cite{shin2024instantdrag, mou2023dragondiffusion, shi2023dragdiffusion, mou2024diffeditor, lu2024regiondrag}, grounding-based approaches (such as bounding boxes)~\cite{chen2023anydoor, xiong2024groundingbooth}, compositing-based algorithms~\cite{alzayer2024magic}, and VAE decoupling methods~\cite{mu2025editable}. While these methods demonstrate capabilities in object manipulation and attribute manipulation, they often struggle to achieve effective element-level editing operations such as addition, translation, scaling, removal, and replacement within a unified framework.

\paragraph{\textbf{Blob-based Controllable Synthesis.}}
Early work established blobs as mid-level primitives for controllable synthesis, primarily in indoor scenes.
BlobGAN~\cite{epstein2022blobgan} first leveraged unsupervised learning to decompose scenes into blobs, enabling layout-level control; 
BlobGAN-3D~\cite{wang2023blobgan} extended this paradigm to 3D, enabling control over camera and 3D object locations. 
Diffusion-based methods further leveraged blob parameters as conditioning signals for text-to-image generation in BlobGEN~\cite{niecompositional}. 
DiffUHaul~\cite{avrahami2024diffuhaul} further employs a training-free procedure to adapt BlobGEN for object dragging in images.
BlobGEN-3D~\cite{liu2024blobgen} formalized blobs as a compositional, 3D-consistent representation to lift 2D scenes into 3D and support free-view synthesis, while BlobGEN-VID~\cite{feng2025blobgen} used blobs as grounding cues for compositional text-to-video generation. 
In contrast, we target image editing rather than generation: we treat blobs as manipulable visual primitives that disentangle layout from appearance, enabling precise element-level operations on existing images with strong identity preservation.

\section{Method}

Sec.~\ref{sec:blob_representation} introduces the blob-related formulations as foundational knowledge. Sec.~\ref{sec:model_architecture} presents the architecture of our model, while Sec.~\ref{sec:self-supervised} and \ref{sec:tailored-strategies} elaborate on the carefully designed training paradigm and strategies tailored for effective learning.

\subsection{Blob-Based Element-level Representation}
\label{sec:blob_representation}
\paragraph{\textbf{Blob Formula}} Fig.~\ref{fig:illustration_blob} illustrates a blob. Geometrically, a blob can be modeled as an ellipse parameterized by $\bm{e}_\tau = [C_x, C_y, a, b, \theta]$, where $(C_x, C_y)$ denote the center coordinates, $a$ and $b$ are the lengths of the semi-minor and semi-major axes, respectively, and $\theta \in [0,\pi)$ is the orientation. Statistically, a blob is modeled as a two-dimensional Gaussian distribution with mean $\bm{\mu} = [\mu_x, \mu_y]$ and covariance $\bm{\Sigma} = \begin{bmatrix} \sigma_{xx} & \sigma_{xy} \\ \sigma_{xy} & \sigma_{yy} \end{bmatrix}$, where $\sigma_{xx}$ and $\sigma_{yy}$ are the variances along the $x$ and $y$ directions, and $\sigma_{xy}$ is the covariance indicating the correlation between $x$ and $y$.

\paragraph{\textbf{Blob Opacity}}
Notably, representing the blob as a Gaussian enables the calculation of opacity across spatial dimensions~\cite{epstein2022blobgan}. In particular, the squared Mahalanobis distance~\cite{mahalanobis1936generalized} to the blob center is computed as:
\begin{equation}
\label{eq:mahalanobis}
\begin{aligned}
    d_M(\bm{x}_{\text{grid}}, \bm{Q}) = (\bm{x}_{\text{grid}} - \bm{\mu})^T \bm{\Sigma}^{-1} (\bm{x}_{\text{grid}} - \bm{\mu}),
\end{aligned}
\end{equation}
where $\bm{x}_{\text{grid}} \in \left\{\left(\frac{w}{W}, \frac{h}{H}\right)\right\}_{w=1..W,\, h=1..H}$ denotes a two-dimensional coordinate map over the image grid, and $\bm{Q}=(\bm{\mu},\bm{\Sigma})$ are the parameters of the blob's bivariate Gaussian. The distance $d_M \in \mathbb{R}^{H\times W}$ is the corresponding distance map that quantifies how far each grid point is from the center $\bm{\mu}$ while accounting for the shape encoded by $\bm{\Sigma}$. Specifically, for each grid index $(w,h)$, \noindent\(d_M[w,h] = \big(\bm{x}_{\text{grid}}[w,h] - \bm{\mu}\big)^T\, \bm{\Sigma}^{-1}\, \big(\bm{x}_{\text{grid}}[w,h] - \bm{\mu}\big).\)
Then, the blob opacity is defined based on this distance:
\begin{equation}
\label{eq:opacity}
\begin{aligned}
    O(\bm{x}_{\text{grid}}) = {\text{sigmoid}}(-d_M),
\end{aligned}
\end{equation}
which maps the distance $d_M$ to values in $(0,1)$. This yields a smooth, center-peaked opacity that gradually decays toward the edges.

\paragraph{\textbf{Blob Composition and Splatting}}
Blob splatting~\cite{epstein2022blobgan} projects $i_{th}$ feature vectors $\bm{f}^i \in \mathbb{R}^{d}$ into a 2D grid with composed blob opacities $O_c^i \in \mathbb{R}^{H\times W}$, producing spatially-aware features $\bm{F_i} \in \mathbb{R}^{H \times W \times d}$. With blobs ordered by depth, the composed opacity, modeling inter-blob occlusion, is
\begin{equation}
\label{eq:composed_opacity}
\begin{aligned}
    O_c^i = O_i \, \odot \, \prod_{j=i+1}^{m} \big(\mathbf{1} - O_j\big),
\end{aligned}
\end{equation}
and per-blob splatting is
\begin{equation}
\label{eq:blob_splatting}
\begin{aligned}
    \bm{F}_i = g_{\text{splatting}} (\bm{f}^i, O_c^i) =  O_c^i \otimes \bm{f}^i ,
\end{aligned}
\end{equation}
where $\odot$ denotes element-wise multiplication on maps and $\mathbf{1}\in\mathbb{R}^{H\times W}$ is the all-ones map. In Eq.~\eqref{eq:blob_splatting}, the map–vector product uses outer-product broadcasting, i.e., $(O_c^i \otimes \bm{f}^i)[h,w,:] = O_c^i[h,w]\, \bm{f}^i$.

\subsection{In-Context Dual Branch Architecture}
\label{sec:model_architecture}
\paragraph{\textbf{Overview.}} Our approach addresses element-level image editing by segmenting the target object as the foreground element and constructing a background through dual masking—removing both the original and target positions of the foreground element. We define foreground as countable "things" (e.g., birds, dogs) and background as uncountable "stuff" regions (e.g., sky, grass), assuming one foreground and one background element per image for simplicity. We design a dual-branch architecture that processes foreground and background separately, where composed opacities $O_c^0$ (background) and $O_c^1$ (foreground) encode their respective layouts. To enhance flexible control over foreground elements, we splat DINOv2~\cite{oquab2023dinov2} features with the foreground opacity $O_c^1$, yielding spatially-aware foreground semantics map $\bm{F}_{1}$. The foreground branch extracts hierarchical features that are progressively fused into the background branch, enabling fine-grained controllable editing through blob representations. Henceforth, we use subscript $1$ for foreground and $0$ for background.

\paragraph{\textbf{Foreground Branch}.} The foreground branch extracts controllable features for injection into the background branch. Let $\operatorname{cat}_0(\cdot)$ denote channel-wise concatenation and $\operatorname{cat}_w(\cdot)$ denote in-context concatenation along the width axis. We construct the inputs as
\begin{equation}
\label{eq:foreground_input_2}
\begin{aligned}
    \bm{C}_{1} &= \operatorname{cat}_0\big(\bm{z}_{1},\; O_{c}^{1},\; \bm{F}_{1}\big) \in \mathbb{R}^{(c+1+d)\times h\times w},\\
    \bm{X}_{1}^t &= \operatorname{cat}_w\Big(\bm{C}_{1},\; \operatorname{cat}_0\big(\bm{z}_t^1,\; O_{c}^{1},\; \bm{F}_{1}\big)\Big) \in \mathbb{R}^{(c+1+d)\times h\times 2w},
\end{aligned}
\end{equation}
where $\bm{z}_{1}\in \mathbb{R}^{c\times h\times w}$ are foreground VAE latents, $O_{c}^{1}\in\mathbb{R}^{1\times h\times w}$ is the foreground composed opacity, $\bm{F}_{1}\in\mathbb{R}^{d\times h\times w}$ is the foreground semantic feature map, and $\bm{z}_t^1\in\mathbb{R}^{c\times h\times w}$ is the noisy foreground latent at timestep $t$.

We use a modified pre-trained diffusion backbone without cross-attention layers to process the foreground input. The input projection layer is modified to handle the dimensionally-changed input $\bm{X}_1^t$. This design leverages pre-trained weights for effective foreground feature processing while focusing solely on visual content.

The foreground branch extracts hierarchical features at multiple resolution levels through the diffusion backbone. For the $i$-th bottleneck block, the extracted features are:
\begin{equation}
\label{eq:foreground_feature_extraction}
\bm{\epsilon}_{\theta}^{i,\text{fg}}(t, \bm{X}_1^t) \in \mathbb{R}^{c_i \times h_i \times w_i},
\end{equation}
where $c_i$, $h_i$, and $w_i$ are the channel, height, and width dimensions at the $i$-th resolution level, respectively. These hierarchical features are progressively injected into the background branch for integration.

\paragraph{\textbf{Background Branch.}} The background branch integrates foreground elements into the scene for controllable generation. Unlike the foreground branch which processes only the foreground region $\bm{z}_t^1$, the background branch operates on the entire image latent $\bm{z}_t$ for proper scene integration. We concatenate the noisy latent $\bm{z}_t$ with reference background conditions $\bm{C}_0$ via in-context format:
\begin{equation}
    \label{eq:background_input_1}
    \begin{aligned}
        \bm{C}_0 &= \operatorname{cat}_0\big(\bm{z}_{0},\; O_{c}^{0}\big) \in \mathbb{R}^{(c+1)\times h\times w},\\
        \bm{X}_0^t &= \operatorname{cat}_w\Big(\bm{C}_0,\; \operatorname{cat}_0\big(\bm{z}_t,\; O_{c}^{0}\big)\Big) \in \mathbb{R}^{(c+1)\times h\times 2w},
    \end{aligned}
\end{equation}
where $\bm{z}_{0}\in \mathbb{R}^{c\times h\times w}$ are background VAE latents and $O_{c}^{0}\in\mathbb{R}^{1\times h\times w}$ is the background composed opacity.

The background branch uses a complete diffusion backbone with cross-attention layers. The input projection layer is modified to handle the dimensionally-changed input $\bm{X}_0^t$. We employ hierarchical feature fusion, progressively injecting foreground features at multiple resolution levels using zero-initialization~\cite{zhang2023adding} $\mathcal{Z}$—initializing the linear layer weights between foreground and background fusion to zero. Feature fusion for the $i$-th block is formulated as:
\begin{equation}
\label{eq:fg_bg_fusion}
\begin{aligned}
    \bm{\epsilon}_{\theta}^{i,\text{enhanced}}(t, \bm{X}_0^t, \bm{X}_1^t) = \bm{\epsilon}_{\theta}^{i,\text{bg}}(t, \bm{X}_0^t) + \omega \cdot \mathcal{Z}(\bm{\epsilon}_{\theta}^{i,\text{fg}}(t, \bm{X}_1^t)),
\end{aligned}
\end{equation}
where $\bm{X}_0^t$ and $\bm{X}_1^t$ are the input conditions for the background and foreground branches, respectively, and $\omega$ is a hyperparameter controlling the fusion strength. For clarity, we omit text-conditioning inputs in the formulation.

\subsection{Self-supervised Training Paradigm}
\label{sec:self-supervised}
\paragraph{\textbf{Disentangle-then-Reconstruct}} 
Obtaining element-level paired supervision for realistic edit operations is challenging and costly. Prior works~\cite{chen2023anydoor, alzayer2024magic} turn to video proxies, which introduce confounds that degrade performance. 
We therefore adopt a self-supervised \emph{Disentangle-then-Reconstruct} paradigm: we treat each existing image as a post-edit result, \emph{disentangle} the foreground element from the background, and construct \emph{dual masks} that remove the element at both a hypothesized pre-edit source and the actual target. We then \emph{reconstruct} by inpainting background at the source and synthesizing the foreground at the target to enforce scene harmony. Concretely, for each image we identify the foreground element's blob in the image as the target (post-edit) state, and sample a synthetic pre-edit blob by randomly perturbing its parameters (center/scale/orientation), upon which we form \emph{dual masks} for source and target, as illustrated in Fig.~\ref{fig:framework}. 
We optimize our model using a noise-prediction objective during training:
\begin{equation}
    \label{eq:diffusion_score_function}
    \begin{aligned}
        \mathcal{L} = \mathbb{E}_{\bm{X}_0^t,\,\bm{X}_1^t,\,\epsilon\sim\mathcal{N}(0, \mathit{I})}\left[\|\epsilon - \epsilon_\theta^{\text{enhanced}}(t, \bm{X}_0^t, \bm{X}_1^t)\|_2^2\right],
    \end{aligned}
\end{equation}
Here, $\bm{X}_0^t$ and $\bm{X}_1^t$ are the background and foreground in-context inputs constructed per Eq.~\eqref{eq:background_input_1} and Eq.~\eqref{eq:foreground_input_2}. This loss drives the model to synthesize the foreground at the target while inpainting the background at the source, ensuring scene harmony.

\paragraph{\textbf{Identity Preservation Loss Function.}}
We impose an identity-preservation loss on the foreground branch to disentangle responsibilities: the foreground branch preserves element-level identity, while the background branch focuses on scene harmonization. During training, the foreground head predicts the noise over masked regions; at inference, we disable this head. Concretely, given the foreground head prediction \(\bm{\epsilon}_{\theta}^{\text{fg}}(t, \bm{X}_1^t)\), the loss is
\begin{equation}
    \label{eq:id_score_function}
    \mathcal{L}_{\text{id}}
    = \mathbb{E}_{\bm{X}_1^t,\,\epsilon\sim\mathcal{N}(0, \mathit{I})}
      \left[\left\| M_1 \odot \big(\epsilon - \bm{\epsilon}_{\theta}^{\text{fg}}(t, \bm{X}_1^t)\big) \right\|_2^2\right],
\end{equation}
where $M_1 \in \{0,1\}^{H\times W}$ is the binary foreground mask, and $\bm{X}_1^t$ is defined in Eq.~\eqref{eq:foreground_input_2}. The overall training objective is
\begin{equation}
    \label{eq:total_loss}
    \mathcal{L}_{\text{total}} = \mathcal{L} + \lambda_{\text{id}}\,\mathcal{L}_{\text{id}},
\end{equation}
where $\lambda_{\text{id}}$ controls the strength of identity preservation. We decay $\lambda_{\text{id}}$ from $1.0$ to $0.6$ over training, which shifts emphasis toward scene harmonization in later stages while retaining identity consistency.

\subsection{Tailored Training Strategies}
\label{sec:tailored-strategies}
\paragraph{\textbf{Random Data Augmentation.}}
To prevent naive copy–paste behavior, we apply extensive augmentations to foreground elements during training, including color jittering, scaling, rotation, random erasing, and perspective transforms. These augmentations (i) compel the model to place foregrounds harmoniously under diverse layouts and appearances, and (ii) strengthen inpainting robustness for incomplete elements. This fosters flexible, context-aware manipulation while maintaining coherence with the background.

\paragraph{\textbf{Random Dropout.}}
With probability $p_{\omega}$ we disable foreground--\allowbreak background fusion by setting $\omega\!\leftarrow\!0$ in Eq.~\eqref{eq:fg_bg_fusion}; with probabilities $p_{\mathsf{feat}}$ and $p_{\mathsf{vae}}$ we set $\bm{F}_{1}\!\leftarrow\!0$ and $\bm{z}_{1}\!\leftarrow\!0$ in Eq.~\eqref{eq:foreground_input_2}. At inference, these hyperparameters can be user-set (e.g., adjust $\omega$ to modulate identity preservation, toggle $\bm{F}_{1}$ or $\bm{z}_{1}$ to trade semantics vs. appearance).


\begin{table*}[t]
  \centering
  \renewcommand{\arraystretch}{1.05}
  \caption{\textbf{Comprehensive comparison of general-purpose methods.} This table quantitatively compares our method against Anydoor~\cite{chen2023anydoor}, GliGEN~\cite{li2023gligen}, and Magic Fixup~\cite{alzayer2024magic} on multiple element-level manipulations. We evaluate identity preservation (CLIP-I/DINO-I), grounding accuracy (MSE), and removal completeness (CLIP-I$^*$/DINO-I$^*$). $\uparrow$ indicates higher is better, while $\downarrow$ indicates lower is better. \label{tab:compare_general_methods} }
  \vspace{-6pt}
  \resizebox{\linewidth}{!}{
  \begin{tabular}{@{}c|ccc|ccc|ccc|ccc|cc@{}}
  \toprule
  \multirow{2}{*}{Method} 
  & \multicolumn{3}{c|}{Addition} 
  & \multicolumn{3}{c|}{Translation} 
  & \multicolumn{3}{c|}{Scaling} 
  & \multicolumn{3}{c|}{Replacement} 
  & \multicolumn{2}{c}{Removal} \\ 
  \cmidrule(l){2-15} 
   & CLIP-I$\uparrow$ & DINO-I$\uparrow$ & MSE$\downarrow$ 
   & CLIP-I$\uparrow$ & DINO-I$\uparrow$ & MSE$\downarrow$ 
   & CLIP-I$\uparrow$ & DINO-I$\uparrow$ & MSE$\downarrow$ 
   & CLIP-I$\uparrow$ & DINO-I$\uparrow$ & MSE$\downarrow$ 
   & CLIP-I$^*\downarrow$ & DINO-I$^*\downarrow$ \\ 
  \midrule
  Anydoor & 86.7 & 81.2 & 6.7 & 85.4 & 81.7 & 6.8 & 83.3 & 83.7 & 9.6 & 81.7 & 80.2 & 9.7 & 39.5 & 13.6 \\ 
  GliGEN   & 70.7 & 57.8 & 6.9 & 71.2 & 62.4 & 7.1 & 78.2 & 69.4 & 9.7 & 68.4 & 60.6 & 9.6 & 40.2 & 15.3 \\ 
  Magic Fixup & 83.7 & 84.5 & 6.6 & 86.0 & 83.7 & 6.8 & 85.5 & 84.2 & 9.2 & 84.5 & 81.2 & 9.2 & 37.2 & 9.7 \\ 
  \midrule
  \rowcolor{gray!10}
  \modelname{}~(Ours) 
  & \textbf{88.3} & \textbf{86.9} & \textbf{6.4} 
  & \textbf{88.9} & \textbf{87.8} & \textbf{6.3} 
  & \textbf{86.5} & \textbf{89.1} & \textbf{8.9} 
  & \textbf{86.2} & \textbf{86.0} & \textbf{9.0} 
  & \textbf{35.3} & \textbf{8.6} \\
  \bottomrule
  \end{tabular}
  }
\end{table*}

\begin{table}[t]
  \centering
  \renewcommand{\arraystretch}{0.92}
  \caption{\textbf{Comparison with point-based dragging methods} on the translation task~\cite{shin2024instantdrag, wu2024draganything, mou2024diffeditor}.  N/A indicates object localization failed, making the metric incomputable.}
  \vspace{-6pt}
  \label{tab:compare_dragging_methods_natural}
  \resizebox{0.6\linewidth}{!}{
  \begin{tabular}{@{}l|ccc@{}}
    \toprule
    Method & CLIP-I$\uparrow$ & DINO-I$\uparrow$ & MSE$\downarrow$ \\
    \midrule
    InstantDrag & 80.6 & 77.7 & N/A \\
    DragAnything & 65.2 & 50.4 & N/A \\
    DiffEditor & 78.7 & 71.8 & 6.9 \\
    \midrule
    \rowcolor{gray!10}
    \modelname{}~(Ours) & \textbf{88.9} & \textbf{87.8} & \textbf{6.3} \\
    \bottomrule
  \end{tabular}}
\end{table}

\section{Experiments}

\label{sec:experiments}
\subsection{Datasets, Benchmark and Metrics}
\paragraph{\textbf{\dataname{} Curation.}}
Building on the BrushData dataset with instance segmentation~\cite{ju2024brushnet}, we curate \dataname{} (1.86M samples) by filtering images and masks, annotating blob parameters, and generating captions. Specifically:
(1) Retain images whose shorter side exceeds 480 pixels.
(2) Keep masks with area ratios in [0.01, 0.9] of the image area and not touching image boundaries.
(3) Fit ellipse parameters to each mask~\footnote{\url{https://docs.opencv.org/4.x/de/d62/tutorial_bounding_rotated_ellipses.html}} and derive 2D Gaussian.
(4) Discard samples with ill-conditioned covariance (below 1e-5).
(5) Generate detailed captions using InternVL-2.5~\cite{chen2024expanding}.

\paragraph{\textbf{\benchmarkname{} Curation.}}
Existing benchmarks~\cite{ruiz2023dreambooth, yang2023paint, lin2014microsoft, zhang2024creatilayout} evaluate either grounding capability or identity preservation, but not both. 
They also do not cover the full spectrum of element-level manipulations (addition, translation, scaling, removal, and replacement). 
To bridge these gaps, we present \benchmarkname{}, a benchmark of 100 curated images evenly spanning the five operation types. 
Each image is annotated with ellipse parameters, a foreground mask, and expert-written detailed descriptions. \benchmarkname{} contains both real-world and AI-generated images across diverse scenarios (indoor and outdoor scenes, animals, landscapes), enabling fair and comprehensive evaluation.

\paragraph{\textbf{Evaluation Metrics.}}
\label{sec:evaluation_metrics}
For \emph{objective evaluation}, we assess:
\begin{itemize}[leftmargin=*]
    \item \emph{Identity Preservation.} We employ CLIP-I~\cite{radford2021learning} and DINO-I~\cite{caron2021emerging} scores to measure the appearance similarity between objects in generated and reference images by extracting and comparing object-level features. For the Removal task, we denote CLIP-I$^*$ and DINO-I$^*$ in the table, with smaller values indicating cleaner removal.
    \item \emph{Grounding Accuracy.} To assess layout control, we extract masks from generated images using SAM~\cite{kirillov2023segment}, fit ellipses to these masks, and measure the Mean Squared Error (MSE) against the ground-truth annotations to quantify spatial accuracy.
    \item \emph{Generation Quality.} We use standard image-quality metrics (FID \cite{fid}, PSNR~\cite{psnr}, SSIM \cite{ssim}, LPIPS~\cite{zhang2018unreasonable}) to assess image quality and harmonization.
\end{itemize} 

For \emph{human evaluation}, we conducted a user study in which 10 participants each assessed 20 result sets. For each metric (fidelity, layout accuracy, and visual harmony), participants selected the single best result among the candidates.

\subsection{Implementation Details.}
\paragraph{\textbf{Training Details.}}
\modelname{} builds on Stable Diffusion v1.5~\cite{ldm}. All images and annotations are resized to $512 \times 512$ pixels. We initialize both foreground and background branches with pretrained UNet weights. The foreground branch is fully fine-tuned, and the background branch is fine-tuned using LoRA~\cite{lora} (rank=64). We use the Adam optimizer~\cite{kingma2014adam} with a learning rate of 1e-5 and weight decay of 0.01. Training is conducted on our curated \dataname{} for 7 days using 24 NVIDIA V100 GPUs with a batch size of 192. To control the fidelity--\allowbreak diversity trade-off, we set dropout probabilities $p_{\omega}$, $p_{\mathsf{feat}}$, and $p_{\mathsf{vae}}$ to 0.1. The identity-preservation loss weight $\lambda_{\text{id}}$ is gradually decayed from 1.0 to 0.6 during training. We use a caption dropout of 0.1 to enable classifier-free guidance at inference.

\paragraph{\textbf{Evaluation Details.}}
\label{sec:eval_details}
We evaluate \modelname{} on the \benchmarkname{} benchmark against six representative open-source baselines. For each editing type, the inputs consist of: (i) the image, including a foreground and the corresponding hole-filled background (for \emph{addition}, we directly provide the clean background and foreground); (ii) blob parameters specifying the initial and target layouts; and (iii) the target foreground element for \emph{addition} and \emph{replacement}.

We categorize the baselines into two groups. \textbf{General Methods} include grounding-based approaches (GliGEN~\cite{li2023gligen}, Anydoor~\cite{chen2023anydoor}) and compositing-based methods (Magic Fixup~\cite{alzayer2024magic}). 
\textbf{Translation-only Methods} consist of point-based dragging approaches for images (DiffEditor~\cite{mou2024diffeditor}, InstantDrag~\cite{shin2024instantdrag}) and videos (
DragAnything~\cite{wu2024draganything}).
These methods perform only positional edits, since only such edits can be easily specified using point annotations. Artifacts in InstantDrag and DragAnything prevent reliable object segmentation; as a result, standard object-level metrics (e.g., DINO-I) are computed on the entire image instead of individual objects, while grounding accuracy, which cannot be meaningfully converted to an image-level metric, is omitted.

\paragraph{\textbf{Baseline Details.}}
For \textbf{Anydoor}, originally designed for mask-guided foreground insertion with harmonization, we adopt a two-pass strategy: (i) inpaint hole-filled backgrounds by feeding them as both foreground and background inputs to obtain a clean background, and (ii) use an operation-specific mask to insert the true foreground at the target location. For \textbf{GliGEN}, whose bounding-box-conditioned insertion cannot handle hole-filled backgrounds, we first recover a clean background via our removal operation, and then insert the foreground at the specified bounding box. For \textbf{Magic Fixup}, a compositing-based harmonization method, we apply rigid transformations to foreground elements according to the editing operation before harmonization. For \textbf{point-based dragging methods}, we use the blob centroids before and after editing as start and end positions to form the dragging points input.

\begin{table}[t]
  \centering
    \caption{\textbf{Comparison of generation quality}~\cite{chen2023anydoor,li2023gligen,alzayer2024magic,wu2024draganything,shin2024instantdrag,mou2024diffeditor}. \label{tab:exp-quality}}   
  \vspace{-6pt}
  \renewcommand{\arraystretch}{0.9} 
  \resizebox{0.95\linewidth}{!}{
    \begin{tabular}{@{}l|l|cccc@{}}
    \toprule
    \textbf{Method} & \textbf{Edit Type} & \textbf{PSNR}$\uparrow$ & \textbf{SSIM}$\uparrow$ & \textbf{LPIPS}$\downarrow$ & \textbf{FID}$\downarrow$ \\
     \midrule
     Anydoor & General & 32.06 & 0.742 & 0.239 & 145.3 \\
     GliGEN  & General & 27.92 & 0.241 & 0.696 & 307.8 \\
     Magic Fixup & General & 31.36 & 0.748 & 0.223 & 131.7 \\
     \rowcolor{gray!10}
     \modelname{}~(Ours) & General & \textbf{32.16} & \textbf{0.751} & \textbf{0.220} & \textbf{102.8} \\
     \midrule
     InstantDrag & Translation Only & 17.38 & 0.680 & 0.185 & 141.9 \\
     DragAnything & Translation Only & 10.75 & 0.282 & 0.611 & 245.7 \\
     DiffEditor & Translation Only & 24.22 & 0.956 & 0.100 & 122.8 \\
     \rowcolor{blue!5}
     \modelname{}~(Ours) & Translation Only & \textbf{29.48} & \textbf{0.975} & \textbf{0.081} & \textbf{74.6} \\
    \bottomrule
    \end{tabular}
   }
\end{table}

\begin{table}[t]
  \centering
    \caption{\textbf{Human evaluation results}~\cite{chen2023anydoor,li2023gligen,alzayer2024magic,wu2024draganything,shin2024instantdrag,mou2024diffeditor}. \label{tab:exp-human}} 
  \vspace{-6pt}
  \renewcommand{\arraystretch}{0.9} 
  \resizebox{0.95\linewidth}{!}{
    \begin{tabular}{@{}l|l|ccc@{}}
    \toprule
    \textbf{Method} & \textbf{Edit Type} & \textbf{Fidelity}$\uparrow$ & \textbf{Layout}$\uparrow$ & \textbf{Harmony}$\uparrow$ \\
     \midrule
     Anydoor & General & 7.5\% & 9.5\% & 8.0\% \\
     GliGEN  & General & 3.0\% & 4.0\% & 5.5\% \\
     Magic Fixup & General & 10.0\% & 11.5\% & 8.0\% \\
     \rowcolor{gray!10}\modelname{}~(Ours) & General & \textbf{79.5\%} & \textbf{75\%} & \textbf{78.5\%} \\
     \midrule
     InstantDrag & Translation Only & 3.0\% & 2.0\% & 1.0\% \\
     DragAnything & Translation Only & 1.5\% & 2.0\% & 1.5\% \\
     DiffEditor & Translation Only & 11.0\% & 13.5\% & 17.5\% \\
     \rowcolor{blue!5}\modelname{}~(Ours) & Translation Only & \textbf{84.5\%} & \textbf{82.5\%} & \textbf{80.0\%} \\
    \bottomrule
    \end{tabular}
   }

\end{table}

\begin{figure*}[ht]
  \centering
    \includegraphics[width=0.82\linewidth]{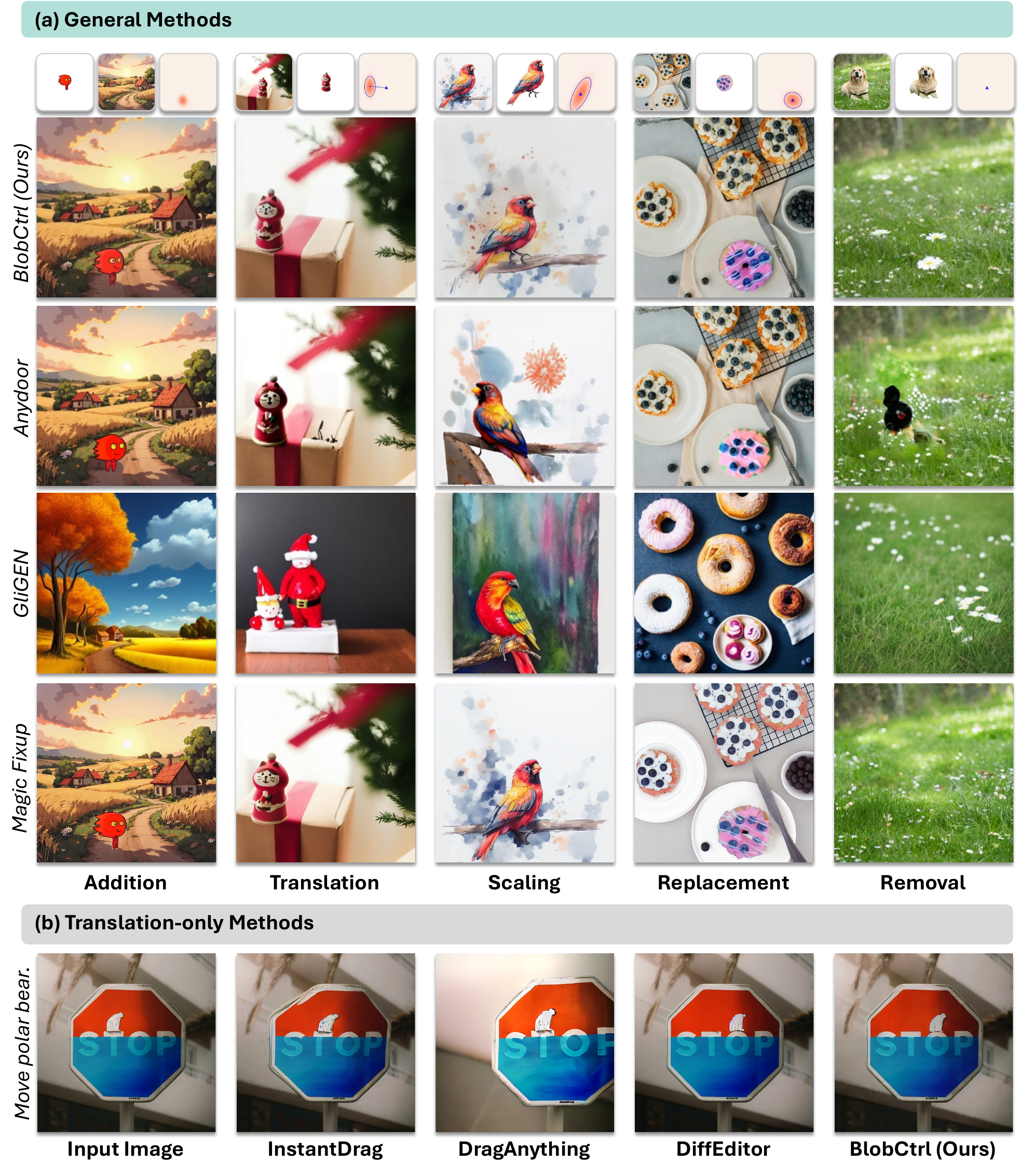}
\vspace{-1em}
\caption{\textbf{Element-level editing comparison across methods.} (a) General Methods supporting diverse element-level operations; (b) Translation-only Methods limited to point-based object relocation. Please zoom in to view source images and manipulation instructions in detail.}
\label{fig:comparisons}
\vspace{-1em}
\end{figure*}

\subsection{Quantitative Evalution}
\paragraph{\textbf{Comparison to State-of-the-Art Methods.}}
As shown in Tab.~\ref{tab:compare_general_methods}, Tab.~\ref{tab:compare_dragging_methods_natural} and Tab.~\ref{tab:exp-quality}, \modelname{} demonstrates consistent and significant improvements over existing methods across all evaluation metrics:

\begin{itemize}[topsep=3.5pt,leftmargin=*]
    \item \textbf{Identity Preservation:} For general methods, \modelname{} achieves substantially higher identity scores on tasks that require preserving elements (addition, translation, scaling, replacement), with average CLIP-I of $87.48$ and DINO-I of $87.45$, outperforming the previous best baseline, Magic Fixup ($84.93$ and $83.40$). For removal tasks, it attains lower CLIP-I$^*$ and DINO-I$^*$ scores (avg. $21.95$ vs. $23.45$), indicating more complete elimination of target elements. In addition, for translation-only tasks, \modelname{} consistently surpasses all drag-based methods.

    \item \textbf{Grounding Accuracy:} \modelname{} demonstrates superior spatial control, achieving a lower average layout MSE than the previous best method, Magic Fixup (${7.65}$ vs. $7.95$), corresponding to a $3.8\%$ relative improvement. This highlights the effectiveness of our blob-based representation for precise element-level manipulation.

    \item \textbf{Generation Quality:} \modelname{} achieves state-of-the-art performance across standard image quality metrics. For general element-level editing, it attains PSNR $32.16$, SSIM $0.751$, LPIPS $0.220$, and FID $102.8$, outperforming all baselines and demonstrating superior global fidelity and realism. For translation-only tasks, our method achieves PSNR $29.48$, SSIM $0.975$, LPIPS $0.031$, and FID $74.6$, consistently surpassing all drag-based methods and highlighting its ability to maintain high-fidelity outputs.
\end{itemize}

We attribute these significant improvements to two key contributions: (1) a high-DoF blob-based representation, enabling precise control over element position, scale, and orientation; and (2) a self-supervised disentangle-then-reconstruct framework, supported by a tailored dual-branch architecture and specialized training strategies, which effectively decouples identity from layout while ensuring robust and harmonious element-level editing.

\paragraph{\textbf{Human Evaluation.}}
The results of Tab.~\ref{tab:exp-human} demonstrate the consistent superiority of \modelname{} across all assessment criteria. For general element-level editing, \modelname{} achieves higher preference rates than existing baselines, with Fidelity $79.5\%$ (vs $10.0\%$), Layout $75.0\%$ (vs $11.5\%$), and Harmony $78.5\%$ (vs $8.0\%$). For translation-only tasks, \modelname{} also outperforms all drag-based methods, achieving Fidelity $84.5\%$, Layout $82.5\%$, and Harmony $80.0\%$.

\subsection{Qualitative Evaluation}
Fig.~\ref{fig:comparisons} shows qualitative comparisons between \modelname{} and state-of-the-art methods. Several consistent observations can be made:

\begin{itemize}[nosep, leftmargin=*]
    \item \textbf{General methods.} GliGEN~\cite{li2023gligen} offers layout control but often breaks identity consistency. Anydoor~\cite{chen2023anydoor} and Magic Fixup~\cite{alzayer2024magic} produce plausible edits but with lower accuracy and visual coherence than ours.
    \item \textbf{Translation-only methods.} InstantDrag~\cite{shin2024instantdrag} fails with large displacements, DragAnything~\cite{wu2024draganything} tends to misinterpret translation as camera motion, and DiffEditor~\cite{mou2024diffeditor} often compromises identity preservation.
\end{itemize}

In contrast, \modelname{} consistently preserves identity, ensures precise layout control, and generalizes well across diverse scenarios while maintaining visual coherence.

\begin{figure*}[!h]
    \centering
    \includegraphics[width=0.96\linewidth]{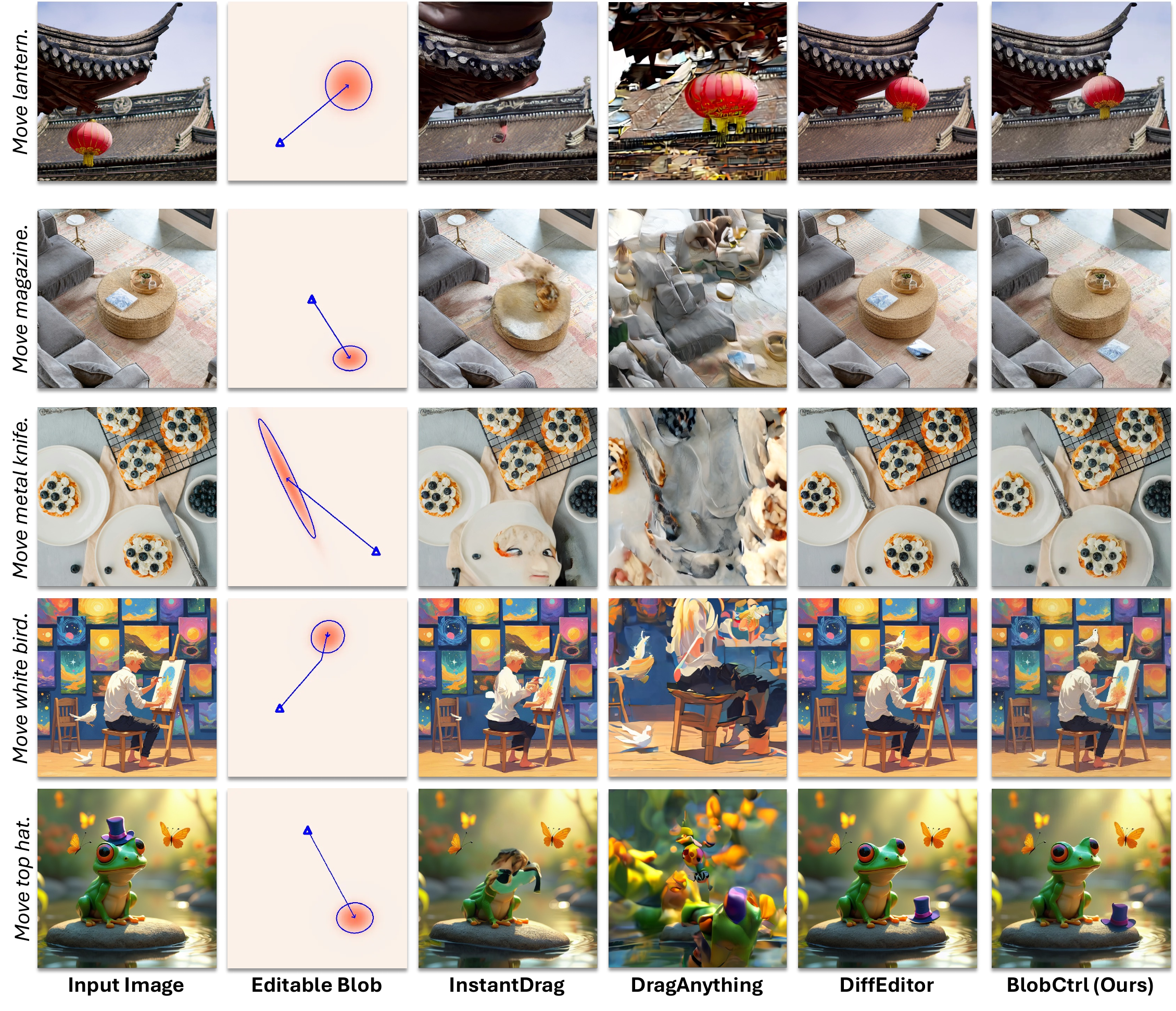}
    \vskip -0.1in
    \caption{\textbf{Additional comparison with translation-only methods.} InstantDrag~\cite{shin2024instantdrag} and DragAnything~\cite{wu2024draganything} fail, while DiffEditor~\cite{mou2024diffeditor} shows lower fidelity compared to our method.}
    \label{fig:additional_translation_only}
\end{figure*}

\begin{figure*}[!h]
    \centering
    \includegraphics[width=0.96\linewidth]{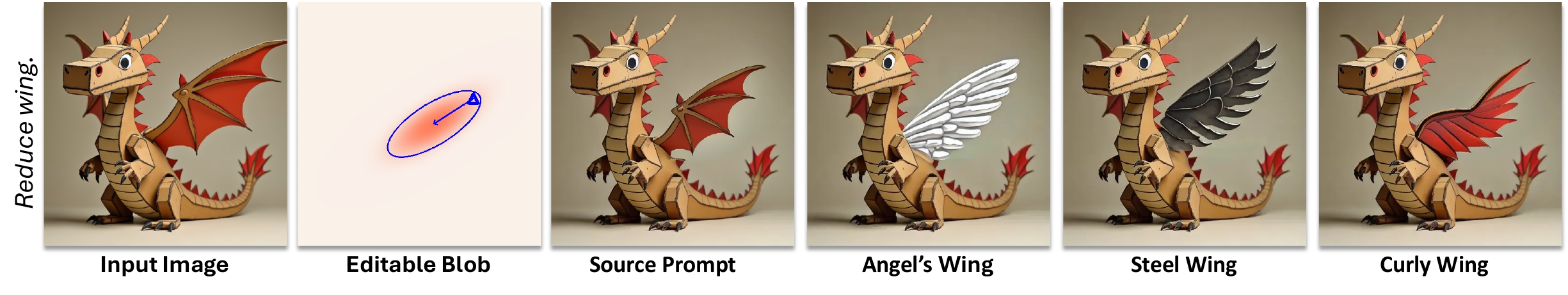}
    \vskip -0.1in
    \caption{\textbf{Results with different text prompts.} The foreground branch is disabled (setting $\omega$ in Eq.~\ref{eq:fg_bg_fusion} to 0), and different prompts guide image edting.}
    \label{fig:additional_text_prompt}
\end{figure*}

\begin{figure*}[!h]
    \centering
    \includegraphics[width=0.82\linewidth]{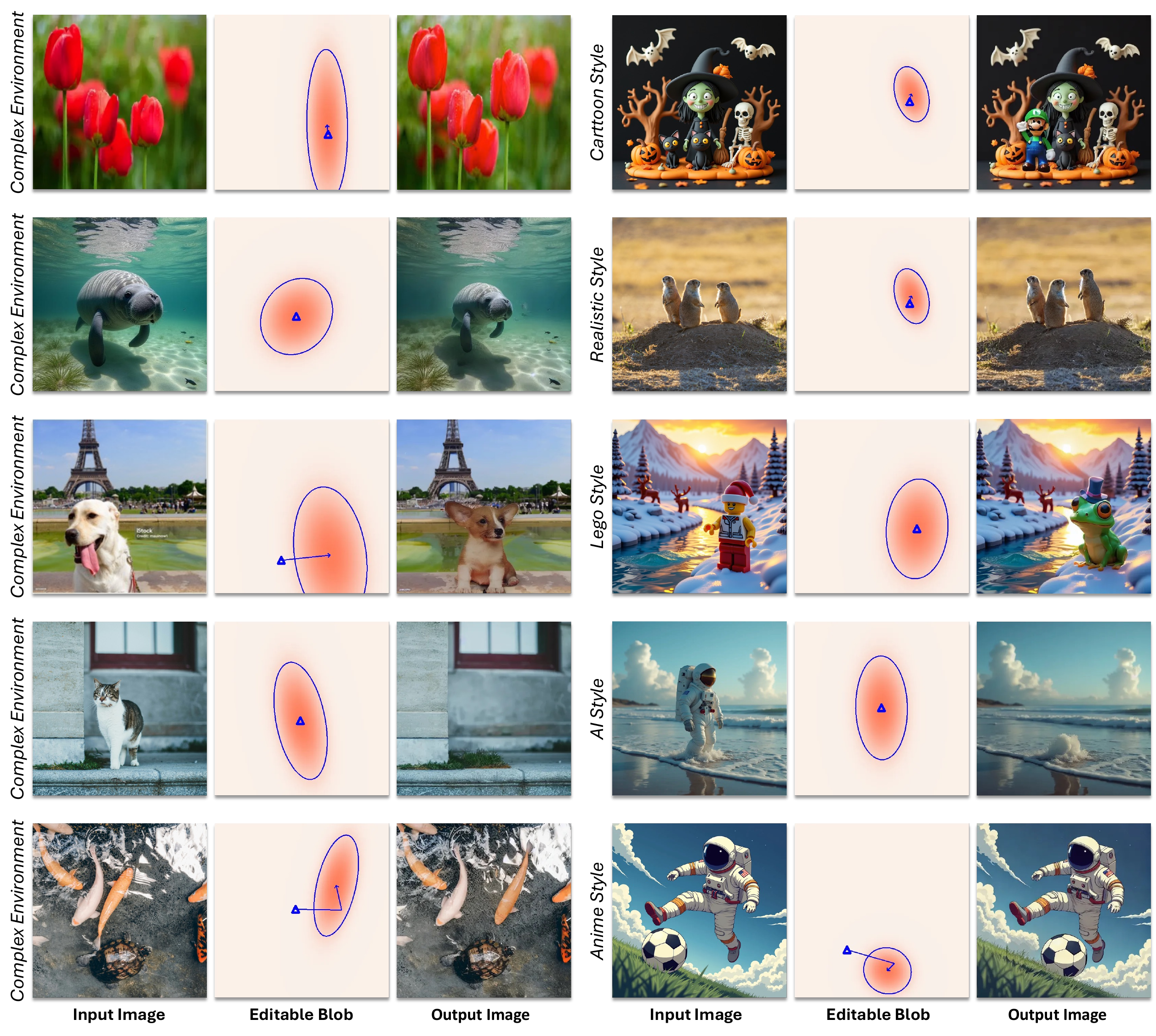}
    \vskip -0.1in
    \caption{\textbf{Editing results under complex settings.} We perform diverse element-level edits—including combined operations such as translation+scale, translation+rotate, and replace+translation—across challenging scenarios (e.g., underwater, crowded scenes, occlusion) and various styles (AI, anime, real-world, LEGO). Our method produces visually plausible results.}
    \label{fig:additional_complex_scene}
\end{figure*}

\begin{figure*}[!h]
    \centering
    \includegraphics[width=0.82\linewidth]{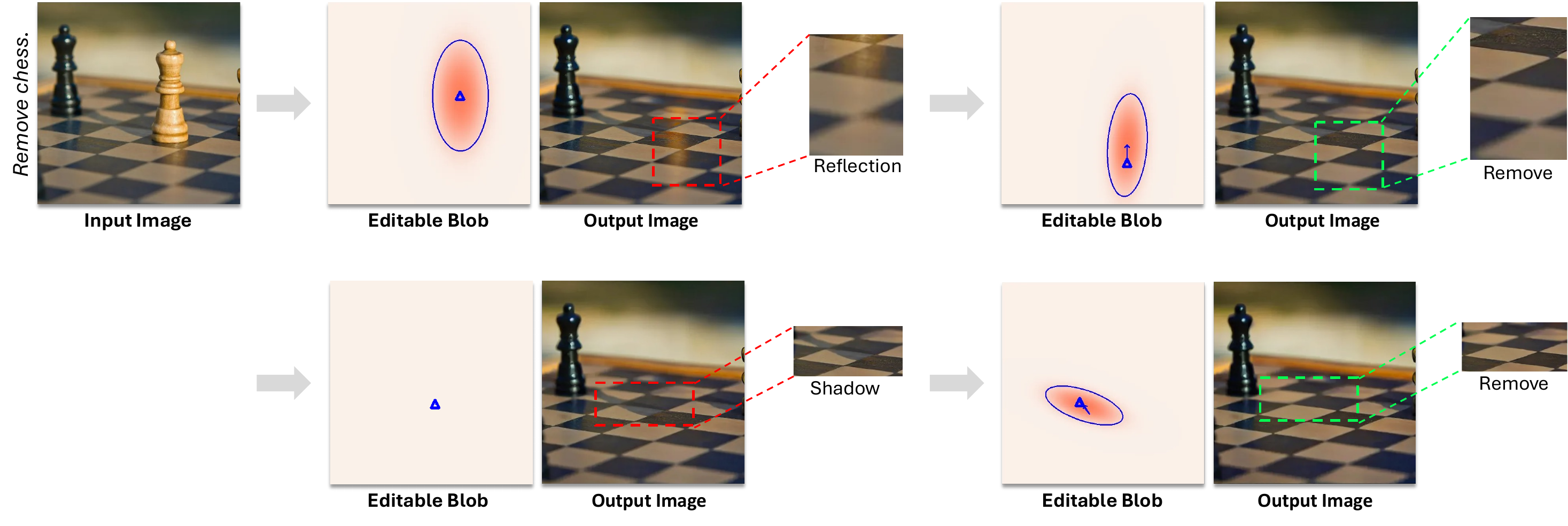}
    \vskip -0.1in
    \caption{\textbf{Results of reflection and shadow removal.} In this setting, shadows and reflections are treated as blob entities and iteratively removed.}
    \label{fig:additional_shadow_effect}
\end{figure*}

Additional comparisons with Translation-only methods are shown in Fig.~\ref{fig:additional_translation_only}, where our approach achieves the best results. Fig.~\ref{fig:additional_text_prompt} illustrates that by adjusting the hyperparameter $\omega$ (Eq.~\ref{eq:fg_bg_fusion}) and the input prompt, our method can switch between reference-followed edits and text-prompt-driven appearance edits. Figures~\ref{fig:additional_complex_scene} and~\ref{fig:additional_shadow_effect} present additional element-level editing results under more complex settings, including diverse edit types (e.g., translate+rotate, replace+scale, translate+scale), challenging scenes (e.g., underwater, crowded scenes, occlusions, shadows, reflections), and varied styles (e.g., real-world, anime, LEGO). Our method produces consistently visually satisfactory results.

\subsection{Ablation Studies}

\begin{figure}[!ht]
    \centering
    \includegraphics[width=1\linewidth]{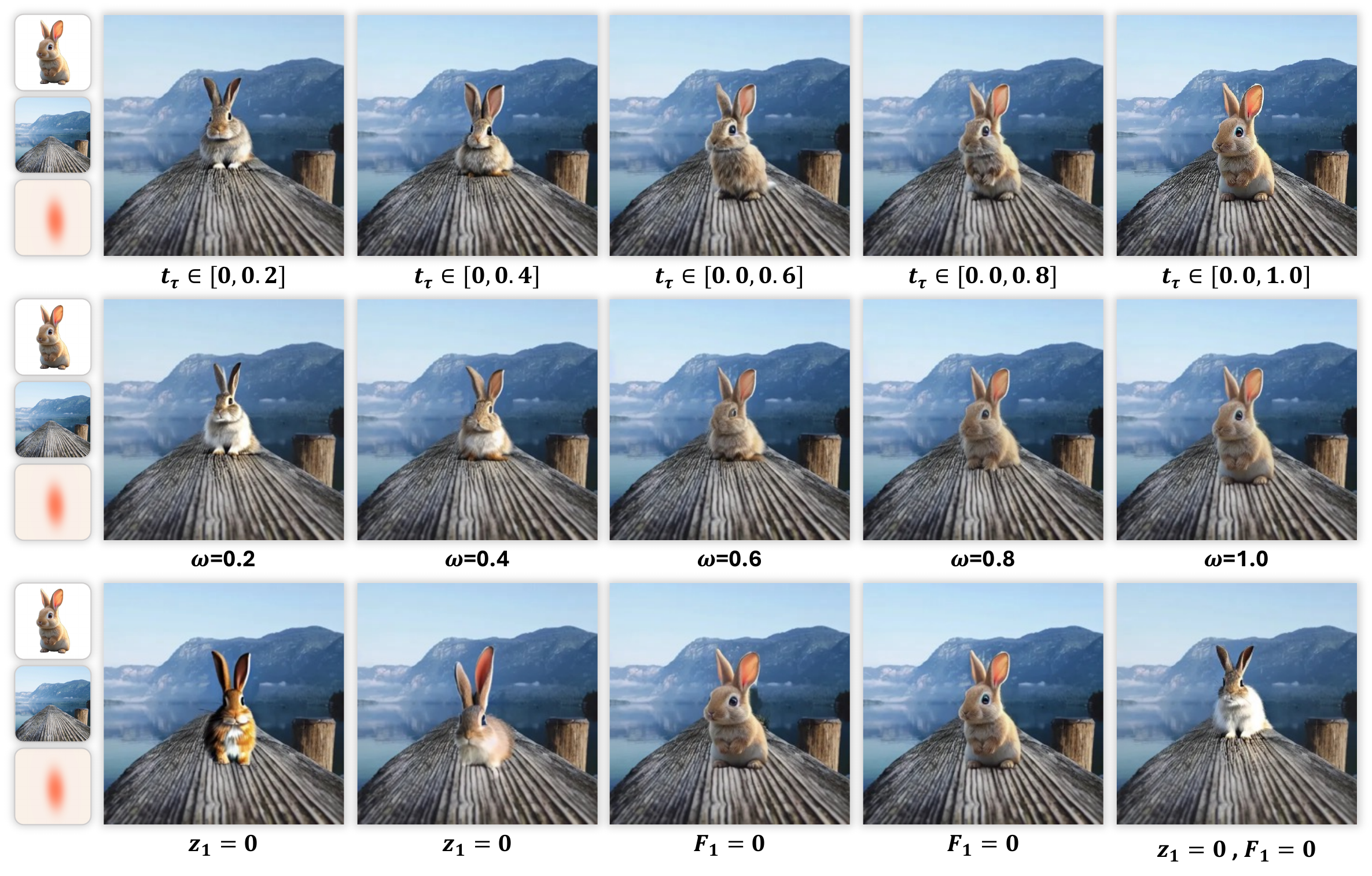}
    \vskip -0.1in
   \caption{\textbf{Foreground–Background Fusion Ablation.} Effect of fusion step ratio $t_\tau$, fusion weight $\omega$, and foreground inputs $\bm{z}_1$, $\bm{F}_1$ on identity preservation and semantic alignment, showing flexible control and diverse outputs.}
    \label{fig:ablation_fusion}
    \vspace{-0.4cm}
\end{figure}

\paragraph{\textbf{Ablation of Foreground–Background Fusion}}
Fig.~\ref{fig:ablation_fusion} presents an ablation study on foreground–background fusion by varying key hyperparameters: fusion weight $\omega$ (Eq.~\eqref{eq:fg_bg_fusion}), fusion step ratio $t_\tau$ (fraction of diffusion steps with foreground–background fusion), and foreground inputs $\bm{z}_{1}$ and $\bm{F}_{1}$ (Eq.~\eqref{eq:foreground_input_2}). Results show that our method enables flexible control over the trade-off between semantic alignment and identity preservation, producing diverse outputs.

\begin{figure}[!ht]
    \centering
    \includegraphics[width=0.92\linewidth]{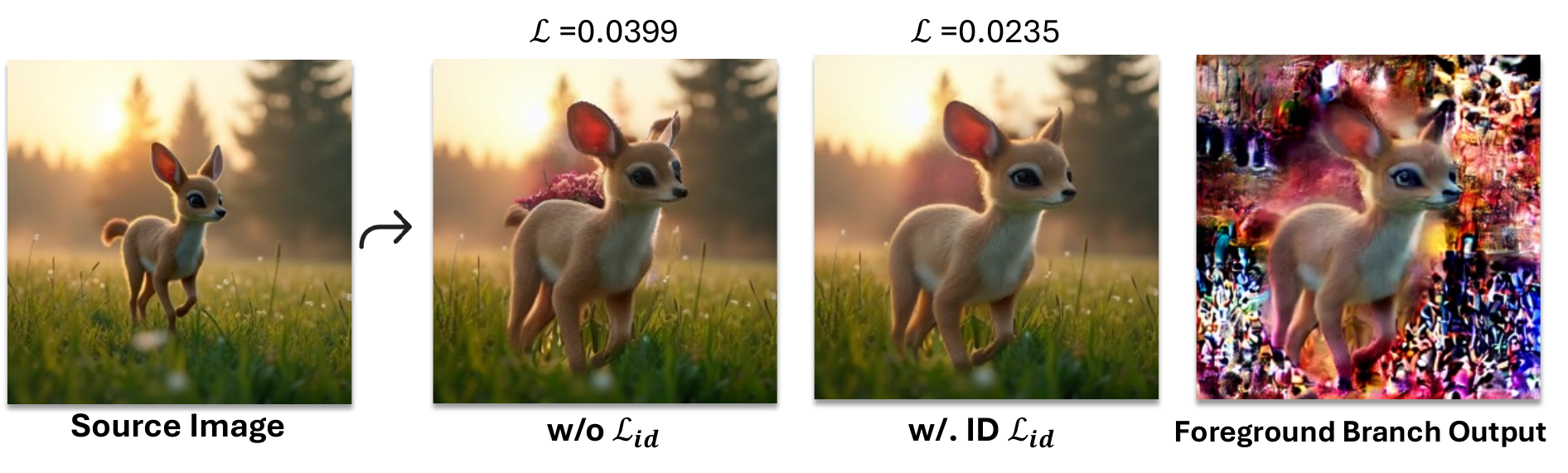}
    \vskip -0.1in
    \caption{\textbf{Ablation of Identity Preservation Loss.} Results of full-image denoising loss and foreground branch outputs.}

    \label{fig:ablation_ID_loss}
    \vspace{-0.4cm}
\end{figure}

\begin{figure}[!ht]
    \centering
    \includegraphics[width=0.86\linewidth]{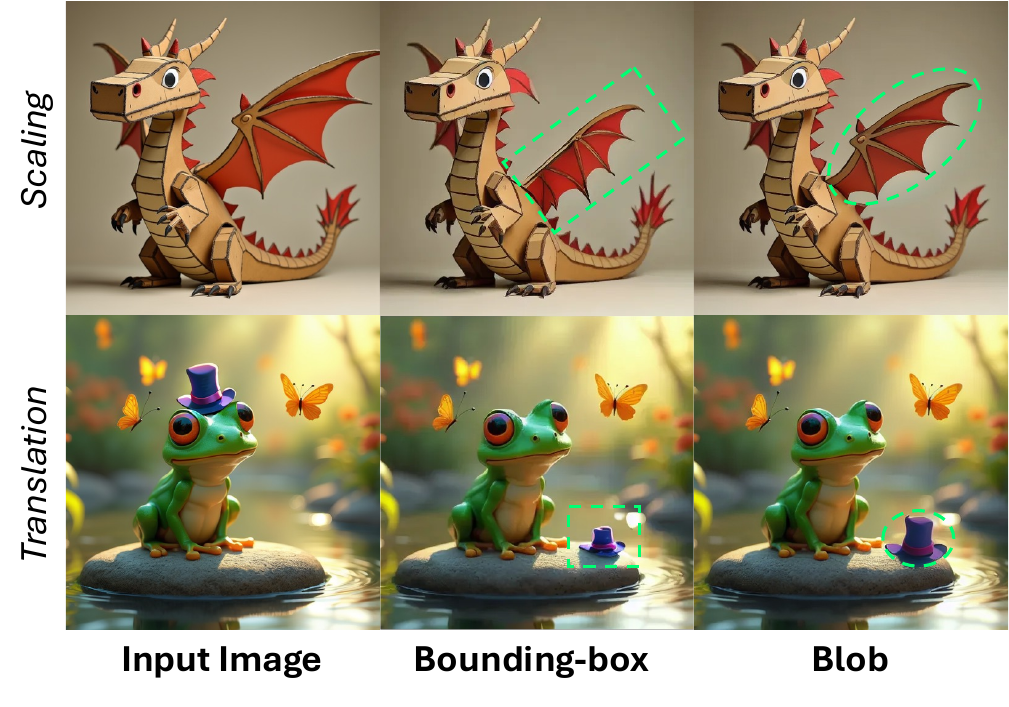}
    \vskip -0.1in
    \caption{\textbf{Ablation on Blob Representation.} Replacing blobs with bounding boxes reduces layout flexibility, while blobs better preserve shapes (e.g., reduced wings) and yield more plausible edits (hat relocation).}
    \label{fig:ablation_blob}
    \vspace{-0.4cm}
\end{figure}

\paragraph{\textbf{Ablation of Identity Preservation Loss Function.}}
Fig.~\ref{fig:ablation_ID_loss} presents an ablation of the Identity Preservation Loss $\lambda_{\text{id}}$ (Eq.~\ref{eq:id_score_function}): without it, the model converges slower (full-image denoising loss 0.0399 vs. 0.0235) and produces lower-quality outputs. 
We additionally decode the foreground branch output corresponding to $\lambda_{\text{id}}$—an output not used during inference. 
This loss acts as a regularizer, encouraging the foreground branch to focus on foreground content and decoupling the functions of the foreground and background branches.

\paragraph{\textbf{Ablation on Blob Representation.}}  
To evaluate their effectiveness, we replace blobs with bounding boxes (Fig.~\ref{fig:ablation_blob}). While bounding boxes are the standard representation for objects, they only have 4-DoF (x, y, w, h), which limits their ability to represent complex shapes. In contrast, blobs have 5-DoF (x, y, a, b, $\theta$), allowing them to better capture irregular shapes and fine details. As a result, our method, which utilizes blobs, offers superior control over object deformation and produces more realistic outcomes (see top of Fig.~\ref{fig:ablation_blob}).

In addition, our blobs are both geometrically and statistically well-defined, interpretable, and interchangeable—taking the form of ellipses geometrically and 2D Gaussian distributions statistically (see Sec.~1 and Sec.~2 of the supplementary materials). This well-defined representation enables smooth and coherent transitions when using blob opacity to represent layouts (Section~\ref{sec:blob_representation}), allowing more precise handling of object details, better preservation of shape, and more natural visual results (see bottom of Fig.~\ref{fig:ablation_blob}.

\section{Limitations and Conclusions}
\label{sec:conclusion}
We present \modelname{}, a flexible framework for element-level editing based on a probabilistic blob representation. Blobs encode spatial information, enabling precise element-level manipulation. With a novel self-supervised dual-branch architecture and customized techniques, \modelname{} achieves consistent edits, controllable flexibility, and state-of-the-art performance on \benchmarkname{}.

Despite its strong capabilities, \modelname{} supports only iterative single-element operations within a single forward pass. Nevertheless, the blob-based representation naturally extends to depth-aware composition, suggesting promising directions for future work.

\begin{acks}
This work is supported by NSFC (No. 62176008), Tencent University Relations (Tencent AI Lab RBFR2024006) and Guangdong Provincial Key Laboratory of Ultra High Definition Immersive Media Technology (Grant No. 2024B1212010006).
\end{acks}

\cleardoublepage

\bibliographystyle{ACM-Reference-Format}
\bibliography{sample-base}

\cleardoublepage
\appendix

\section{Gaussian to Ellipse Conversion}

A 2D Gaussian distribution is defined by its mean $\boldsymbol{\mu} = (\mu_x, \mu_y)$ and covariance matrix $\boldsymbol{\Sigma}$:

\begin{equation}
    \boldsymbol{\Sigma} = \begin{bmatrix} 
        \sigma_x^2 & \rho\sigma_x\sigma_y \\
        \rho\sigma_x\sigma_y & \sigma_y^2
    \end{bmatrix}.
\end{equation}

The level sets of this distribution are ellipses. For a confidence level $\alpha$, the corresponding \emph{confidence ellipse} is given by:

\begin{equation}
    (\mathbf{x}-\boldsymbol{\mu})^T \boldsymbol{\Sigma}^{-1} (\mathbf{x}-\boldsymbol{\mu}) = \chi^2_2(\alpha),
\end{equation}

where $\chi^2_2(\alpha)$ is the upper $\alpha$-quantile of the chi-square distribution with 2 degrees of freedom. The semi-major and semi-minor axes of the ellipse are proportional to the square root of the eigenvalues of $\boldsymbol{\Sigma}$ multiplied by $\sqrt{\chi^2_2(\alpha)}$, and the rotation angle is determined by the eigenvectors.

\section{Ellipse to Gaussian Conversion}

Conversely, given an ellipse with center $(h,k)$, semi-major axis $a$, semi-minor axis $b$, and rotation angle $\theta$ (corresponding to a confidence level $\alpha$), the Gaussian distribution can be reconstructed as:

\begin{equation}
    \boldsymbol{\mu} = \begin{pmatrix} h \\ k \end{pmatrix}, \quad
    \boldsymbol{\Sigma} = \frac{1}{\chi^2_2(\alpha)} \mathbf{R}(\theta) 
    \begin{bmatrix} a^2 & 0 \\ 0 & b^2 \end{bmatrix} 
    \mathbf{R}(\theta)^T,
\end{equation}

with the rotation matrix defined by

\begin{equation}
    \mathbf{R}(\theta) = 
    \begin{bmatrix}
        \cos\theta & -\sin\theta \\
        \sin\theta & \cos\theta
    \end{bmatrix}.
\end{equation}

This relationship allows a precise mapping between probabilistic blob representations and geometric ellipse controls, taking into account both the confidence level and the orientation of the ellipse.

\section{Justification of Baseline Selection}
In Sec. 4, we compare our approach against six representative baselines. Specifically, we include three methods capable of handling multiple types of element-level editing: 
\begin{enumerate}
    \item GliGEN~\cite{li2023gligen}, a method that specifies layouts using bounding boxes.
    \item Anydoor~\cite{chen2023anydoor}, a method that specifies layouts using segmentation.
    \item Magic Fixup~\cite{chen2023anydoor}, a method based on compositing and harmonization.
\end{enumerate}
as well as three methods restricted to translation-based editing:
\begin{enumerate}
    \item DiffEditor~\cite{mou2024diffeditor}, a point-based dragging method that designs different diffusion sampling algorithms for each type of edit.
    \item InstantDrag~\cite{shin2024instantdrag}, a point-based dragging method that predicts sparse optical flow from drags and uses it to guide the editing.
    \item DragAnything~\cite{wu2024draganything}, a point-based dragging method that represents objects using segmentation. This method was originally developed for motion-controllable video generation, and we use the final frame as the edited output.
\end{enumerate}

We exclude several other methods for the following reasons:
\begin{enumerate}
    \item DiffUHaul~\cite{avrahami2024diffuhaul} has not been released.
    \item Image Sculpting~\cite{yenphraphai2024image} relies on per-image optimization rather than generalizable editing.
    \item DragonDiffusion~\cite{mou2023dragondiffusion} is an earlier version of DiffEditor~\cite{mou2024diffeditor}.
    \item RegionDrag~\cite{lu2024regiondrag}, published earlier than both InstantDrag~\cite{shin2024instantdrag} and DiffEditor~\cite{mou2024diffeditor}, is a point-based image dragging method similar to these two approaches.
    \item ObjectStitch~\cite{song2023objectstitch}, published earlier than Magic Fixup~\cite{chen2023anydoor}, and IMPRINT~\cite{song2024imprint}, published around the same time as Magic Fixup, are both similar to Magic Fixup, being methods based on compositing and harmonization.
    \item Image Sculpting~\cite{yenphraphai2024image} involves a complex process that requires manual adjustments for each sample during editing.
    
\end{enumerate}

Taken together, the six selected baselines encompass a range of approaches, including point-based dragging, grounding-based methods, compositing techniques, and even a motion-controllable video generation model. This diverse set provides a comprehensive foundation for evaluating our method.

\section{Limitations of Methods Without Multiple Task Support}

Among the baselines introduced in the previous section, some methods do not support multiple types of generative editing tasks:

Additionally, several other methods have their own limitations:
\begin{enumerate}
    \item Point-based dragging methods like DiffEditor~\cite{mou2024diffeditor}, RegionDrag~\cite{lu2024regiondrag}, InstantDrag~\cite{shin2024instantdrag}, and DragonDiffusion~\cite{mou2023dragondiffusion} are constrained to translation-based editing due to their reliance on sparse point trajectories as input. While interpolation between start and end points is possible, these methods cannot handle more complex operations such as object addition, removal, scaling, or replacement. For example, scaling requires defining points in multiple directions, not just a single direction.
    \item DiffUHaul~\cite{avrahami2024diffuhaul} is a training-free approach that supports only translation-based operations and has not been publicly released.
    \item Image Sculpting~\cite{yenphraphai2024image} is a 3D perception-based method designed for editing meshes in 3D space. While it supports various operations, the process is complex and requires per-sample reconstruction, along with specialized software like Blender for mesh editing.
    \item ObjectStitch~\cite{song2023objectstitch} and IMPRINT~\cite{song2024imprint} focus on compositing and harmonizing foreground and background, but do not explicitly support translation, removal, or scaling operations.
\end{enumerate}

These limitations highlight the advantages of our method, which supports a broader range of generative editing tasks, offering greater flexibility and control over the final output.

\begin{figure*}[ht]
  \centering
    \includegraphics[width=\linewidth]{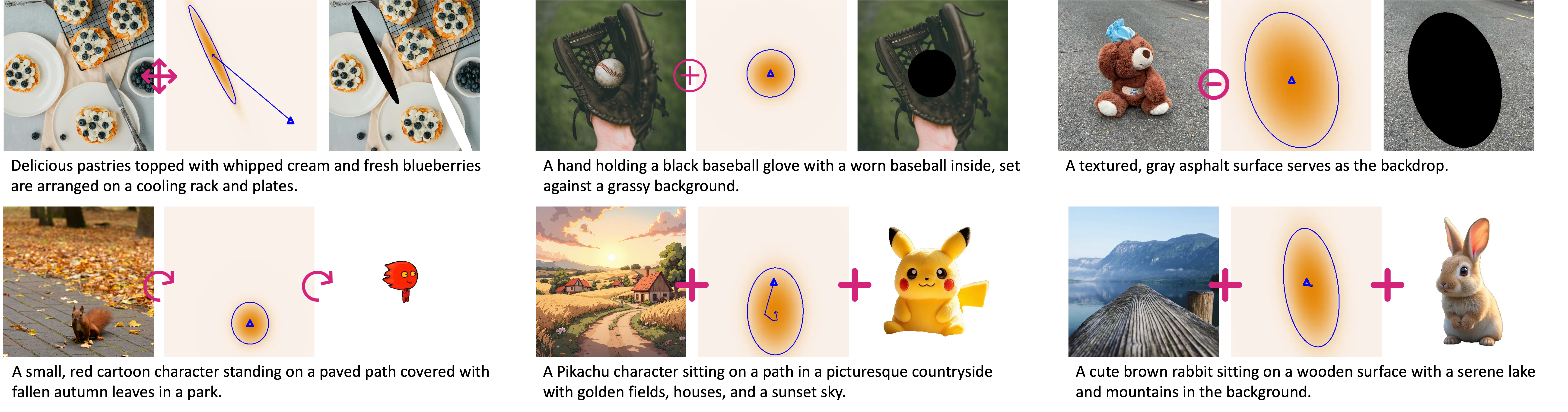}
 \caption{Overview of the BlobBench.}
    \label{fig:blobbench}
\end{figure*}

\begin{figure*}[ht]
  \centering
    \includegraphics[width=0.86\linewidth]{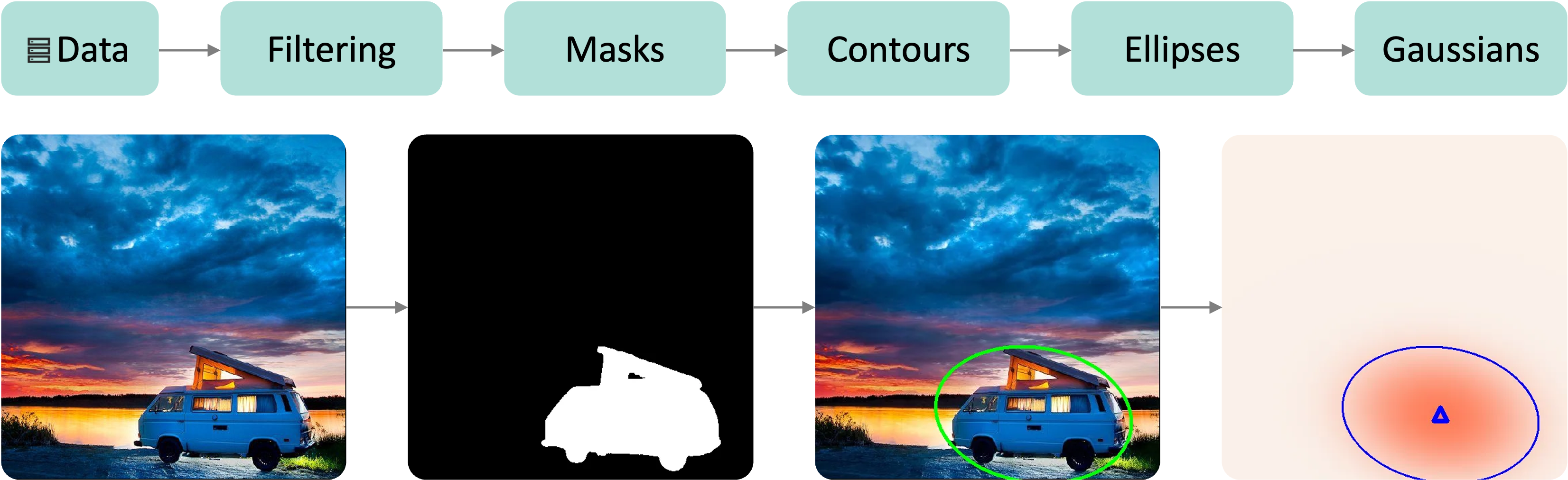}
 \caption{The BlobData curation workflow.}
    \label{fig:blobdata}
\end{figure*}

\section{BlobBench and BlobData}

\emph{BlobBench} is a comprehensive benchmark consisting of 100 curated images, evenly distributed across various element-level operations, including addition, translation, scaling, removal, and replacement. Each image is annotated with ellipse parameters, foreground masks, and textual descriptions, incorporating both real-world and AI-generated images from diverse scenarios such as indoor/outdoor environments, animals, and landscapes (see Fig.~\ref{fig:blobbench}). 

In parallel, \emph{BlobData} is a large-scale dataset comprising 1.86 million samples sourced from BrushData~\cite{ju2024brushnet}. The curation process involves several key steps:

\begin{itemize}
    \item \textbf{Image Filtering.} The source images are filtered to retain those with a minimum short side length of 480 pixels, valid instance segmentation masks, and masks with area ratios between 0.01 and 0.9 of the total image area. Masks touching image boundaries are excluded.
    \item \textbf{Parameter Extraction.} Ellipse parameters are extracted using OpenCV's ellipse fitting algorithm, followed by the derivation of corresponding 2D Gaussian distributions. Invalid samples with covariance values below 1e-5 are removed.
    \item \textbf{Annotation.} Detailed textual descriptions for each image are generated using InternVL-2.5~\cite{chen2024expanding}, providing rich contextual information for each sample.
\end{itemize}

This curated dataset, combining detailed annotations and a diverse set of real-world and synthetic images, serves as the foundation for diverse element-level operations.

\end{document}